%% file: root.tex
\documentclass[letterpaper, 10 pt, conference]{ieeeconf}  

\IEEEoverridecommandlockouts                              

\usepackage[usenames,dvipsnames]{xcolor}

\newcommand{\adp}[1]{\textcolor{Red}{[Andrea: \textbf{#1}]}}

\usepackage{tikz}
\usetikzlibrary{shapes.geometric, arrows.meta, positioning}
\usepackage{float} 
\usepackage{optidef}
\usepackage[linesnumbered]{algorithm2e}\RestyleAlgo{ruled}
\usepackage{comment}
\usepackage{tabularx}
\usepackage{multirow}
\usepackage{array}
\renewcommand{\arraystretch}{1.4}   
\usepackage{amssymb}
\usepackage{pgf}
\usepackage{graphicx}
\usepackage{hyperref}

\usepackage{pgf}
\usepackage{lmodern}

\usepackage{pgfplots}
\pgfplotsset{compat=1.18}

\usepackage{tikz}
\usetikzlibrary{arrows.meta}
\usetikzlibrary{fadings}
\usetikzlibrary{colorbrewer}

\usepackage{makecell}

\usepackage{transparent}

\usepackage{pdfcol}

\title{\LARGE \bf
CACTO-BIC: Scalable Actor-Critic Learning via Biased Sampling and GPU-Accelerated Trajectory Optimization
}

\author{Elisa Alboni$^{1}$, Pietro Noah Crestaz$^{1,2}$, Elias Fontanari$^{1}$, Andrea Del Prete$^{1}$
\thanks{$^{1}$ are with the Dept. of Industrial Engineering, University of Trento, Italy        
[{\tt\footnotesize elisa.alboni},{\tt\footnotesize pietronoah.crestaz},{\tt\footnotesize andrea.delprete}]\tt\footnotesize@unitn.it}
\thanks{$^{2}$ is with LAAS-CNRS, Université de Toulouse, CNRS, Toulouse, France. \tt\footnotesize pncrestaz@laas.fr}
}

\begin{document}

\maketitle
\thispagestyle{empty}
\pagestyle{empty}

\begin{abstract}

Trajectory Optimization (TO) and Reinforcement Learning (RL) offer complementary strengths for solving optimal control problems. TO efficiently computes locally optimal solutions but can struggle with non-convexity, while RL is more robust to non-convexity at the cost of significantly higher computational demands. CACTO (Continuous Actor-Critic with Trajectory Optimization) was introduced to combine these advantages by learning a warm-start policy that guides the TO solver towards low-cost trajectories. However, scalability remains a key limitation, as increasing system complexity significantly raises the computational cost of TO. 

This work introduces CACTO-BIC to address these challenges. CACTO-BIC improves data efficiency by biasing initial-state sampling leveraging a property of the value function associated with locally optimal policies; moreover, it reduces computation time by exploiting GPU acceleration. Empirical evaluations show improved sample efficiency and faster computation compared to CACTO. Comparisons with PPO demonstrate that our approach can achieve similar solutions in less time. Finally, experiments on the AlienGO quadruped robot demonstrate that CACTO-BIC can scale to high-dimensional systems and is suitable for real-time applications.

\end{abstract}

\section{INTRODUCTION}
Trajectory Optimization (TO) is a widely used and flexible technique for solving robotic control problems. In TO, the high-level task is formulated as a constrained Optimal Control Problem (OCP), where the optimization variables are the system's state and control trajectories. Constraints enforce compliance with system dynamics and kinematics, actuator limits, and task-specific requirements. However, OCPs are typically highly non-convex, making gradient-based solvers prone to converge to poor local minima. While global methods based on the Hamilton–Jacobi–Bellman equation or Dynamic Programming~\cite{bellman1954theory} exist, their applicability is limited by the curse of dimensionality.

Deep Reinforcement Learning (RL) emerged as an alternative framework, particularly for continuous state and action spaces. Algorithms such as DDPG \cite{DDPG}, SAC \cite{SAC}, and PPO \cite{PPO} have demonstrated strong performance in robotic control tasks. Due to their exploratory nature, RL methods are generally less sensitive to local minima but they typically suffer from high sample complexity and long training times.

To overcome the complementary limitations of TO and RL, hybrid approaches combining the two have recently gained significant research attention. 
A popular choice is to rely on \textit{TO imitation}: policies are trained to mimic TO or model predictive control (MPC) solutions---through value-based or action-based imitation---to reduce online computational costs and to leverage sensor feedback \cite{carius2020mpc, ghezzi2023imitation}. 
However, these methods neither improve TO solution's quality, nor guarantee constraint satisfaction. 
Accounting for policy approximation errors in the TO problem can lead to better results \cite{levine2013guided, lidec2022enforcing}, but it inherits the same limitations. 

Other methods use learned policies or value functions to warm-start TO or to define terminal costs, thereby accelerating optimization and guiding the solver toward improved solutions \cite{reiter2024ac4mpc, ceder2024bird}. While effective for TO, these approaches provide no benefits for RL training.

An increasingly prominent class of approaches embeds TO directly within the RL framework to improve training efficiency. 
These methods can be classified based on where TO is included: \textit{post-policy}, \textit{pre-policy}, or as a \textit{residual} policy. 
\textit{Post-policy} methods evaluate TO after the policy.
Some methods learn cost or constraint parameters \cite{romero2024actor, zarrouki2024safe}, 
leveraging MPC's safety and stability guarantees, but require solving TO online and face convergence challenges. 
Others use TO as actor and learn its terminal cost \cite{lowrey2018plan, jordana2025infinite}, 
accelerating training and improving constraint handling, yet potentially yielding suboptimal solutions or neglecting sensor feedback. Another variant initializes TO with an RL policy \cite{CACTO,morgan2021model}, improving convergence speed and solution quality, yet still does not exploit sensor feedback. 
In \textit{pre-policy} methods, TO generates reference trajectories or auxiliary information that are inputted to the RL policy, effectively speeding up training \cite{jenelten2024dtc}. 
\textit{Residual} methods instead learn a residual policy to improve TO-generated control inputs with learned corrections~\cite{silver2018residual}.
While both \textit{pre-policy} and \textit{residual} methods accelerate training, they require online TO and depend on its ability to find high-quality solutions; moreover, learned policies may violate constraints even when TO does not. \\

In this work, we extend CACTO~\cite{CACTO,alboni2024cacto}, a \textit{post-policy} algorithm that exploits the interplay between TO and RL to accelerate training. The actor policy generates the initial guess for TO, leveraging the exploratory nature of RL to avoid convergence to poor local minima, while TO guides the learning process of the RL agent. 

The primary limitation of CACTO is scalability. As the system complexity increases, TO becomes more expensive, and actor–critic training requires more iterations to converge.
We investigate strategies to reduce the computational burden in both TO and RL phases. Our main contributions are:
\begin{itemize}
    \item A method to identify state-space regions where an improvement of the actor policy is more likely.
    \item A new version of CACTO's algorithm, called CACTO-BIC, that exploits biased initial conditions (BIC) and GPU-based computation to achieve improved sample and time efficiency.
    \item A JAX-based~\cite{jax2018github} open-source implementation of CACTO-BIC that exploits GPUs to solve TO problems and train neural networks.
    \item The first validation of CACTO on real hardware through experiments on a quadruped robot.
\end{itemize}

\section{BACKGROUND}
This section summarizes the latest formulation of CACTO (originally called CACTO-SL~\cite{alboni2024cacto}), an algorithm for solving finite-horizon discrete-time optimal control problems such as:
\begin{mini}|l|[2]<b>
{\scriptstyle{X, U}}
{L(X,U) \triangleq \sum_{k=0}^{T -1} l_k(x_k,u_k) + l_T(x_T)
}
{\label{OCP_optimizationProblem}}{}
\addConstraint{x_{k+1}}{= f_k(x_k,u_k), \quad k=0,\dots,T-1}
\addConstraint{|u_k|}{\le u_{\max}, \quad\quad\quad\,\,\,\, k=0,\dots,T-1}
\addConstraint{x_0}{= x_{\mathrm{init}},}
\end{mini}
where $X = \{x_0, \dots, x_T\}$ and $U = \{u_0, \dots, u_{T-1}\}$ are the state and control sequences, with $x_k \in \mathbb{R}^n$ and $u_k \in \mathbb{R}^m$. The cost $L(\cdot)$ combines running costs $l_k$ and a terminal cost $l_T$, while the constraints enforce system dynamics, control bounds, and initial conditions. 

CACTO begins by solving $N$ TO problems from randomly sampled initial states, using standard warm-starting (e.g., setting all states to $x_{\mathrm{init}}$ and controls to zero). As the OCP horizon is finite, the actor policy and the critic value are time dependent, so also the initial time is randomized in each TO instance and the time is appended to the state vector, $\tilde{x}=[x,t]$. For each state of each optimized trajectory, CACTO computes the partial $K$-step cost-to-go $\bar{V}$, where $K$ is a user-defined parameter representing the Temporal Difference lookahead horizon. The value function's gradient $\bar{V}_x$ is also computed, using the backward pass of Differential Dynamic Programming (DDP)~\cite{jacobson1970differential}. These values $(\bar{V}, \bar{V}_x)$, together with the associated state, control, and state after $K$ steps, are stored in a replay buffer.

Afterwards, the critic and actor neural networks are trained for $M$ iterations using mini-batches sampled from the replay buffer. The critic approximates the value function and its gradient by matching $\bar{V}$ and $\bar{V}_x$:
\begin{mini!}|l|[2]<b>
{\scriptstyle{\theta^V}}
{
(\bar{V} - V(\tilde{x} | \theta^V))^2 + 
k_s (\bar{V}_x - S_x V_{\tilde{x}} (\tilde{x}|\theta^V))^2
}
{\label{critic_update}}{}
\end{mini!} 
where $V(\cdot)$, $V_{\tilde{x}}(\cdot)$ 
and $\theta^V$ are the critic network, its gradient (with respect to $\tilde{x}$) and its parameters, $k_s$ is the weight of the gradient term, and $S_x$ the selection matrix to exclude the partial derivative of $V$ with respect to time. 
The actor is updated by minimizing the Q-value: 
\begin{mini!}|l|[2]<b>
{\scriptstyle{\theta^\mu}}
{l(\tilde{x},\mu(\tilde{x}|\theta^\mu))+V_{\theta^V}(f(\tilde{x},\mu(\tilde{x}|\theta^\mu)))}
{\label{actor_update}}{}
\end{mini!}
where $\mu(\cdot)$ and $\theta^\mu$ are the actor network and its parameters. 
The improved actor policy then generates rollouts that warm-start the subsequent TO problems, closing the loop between TO and RL. 

In~\cite{CACTO}, CACTO 
was shown to outperform other RL algorithms such as DDPG~\cite{DDPG} and PPO~\cite{PPO} as warm-start provider in terms of training time.

\section{BIASED INITIAL STATES SAMPLING}
Identifying regions of the state space with high potential for policy improvement is a challenging problem.  

Exploration strategies in RL can be broadly categorized into undirected and directed approaches. Undirected exploration relies on stochastic action selection, such as $\epsilon$-greedy policies, to explore the state space without explicitly accounting for uncertainty or novelty. Directed exploration, in contrast, uses signals derived from the agent’s learning process or auxiliary models to guide behavior toward less familiar or more uncertain regions. A common class of directed methods employs intrinsic rewards or exploration bonuses, including count-based or approximate count methods \cite{bellemare2016unifying}, uncertainty metrics \cite{lowrey2018plan}, and curiosity-driven approaches such as prediction-error bonuses and random network distillation \cite{pathak2017curiosity,burda2018exploration}. Several works have also explored initial-state or restart-based exploration. These approaches often assign scores to states to prioritize which ones to explore further. Criteria for scoring include the system’s sensitivity~\cite{parsa2023where2start}, the familiarity~\cite{schenke2021improved}, the uncertainty~\cite{yin2023sample}, or the TD error \cite{tavakoli2018exploring}. \cite{messikommer2024contrastive}, proposes a structured replay buffer that groups states by task relevance and prioritizes sampling from unmastered sub-tasks.
When available, prior or expert knowledge can also be leveraged to further guide exploration. \\

We address the exploration problem by leveraging the insight that the value function $\bar{V}(x)$ associated with locally optimal solutions is generally \textit{piecewise continuous}. 

\subsection{Motivating Example: Discontinuous Value Function}
\label{sec:1Dex}
\begin{figure}[tbp]
    \centering
    \includegraphics[width=0.9\columnwidth]{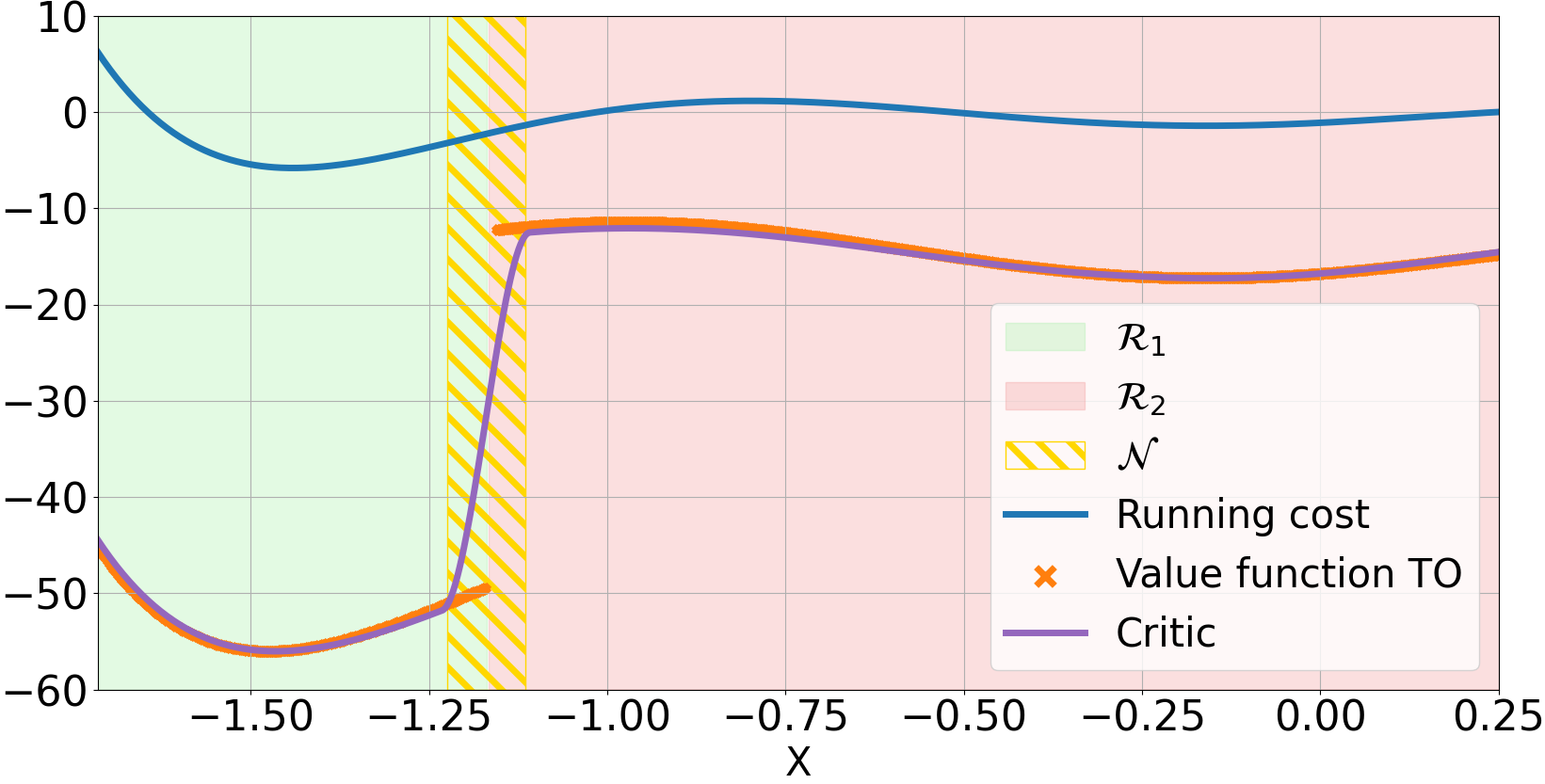}
    \caption{Cost and Value obtained with TO using a naive initial guess. The critic smooths the Value's discontinuities.}
    \label{fig:f1}
\end{figure}

Let us illustrate how the structure of the value function associated with a locally optimal policy can help us tackle the exploration problem.
We focus on a toy problem with a 1D state, single integrator dynamics, and a cost with two local minima (see Fig.~\ref{fig:f1}). 
Solving TO problems with a naive initial guess, highlights the presence of two basins of attraction, $\mathcal{R}_1$, and $\mathcal{R}_2$, each corresponding to a different local minimum (see Fig.~\ref{fig:f1}). 
Within each basin, the real Value $\bar{V}$ is continuous, but it is discontinuous at the shared boundary $\mathcal{R}_1 \cap \mathcal{R}_2 \triangleq \partial \mathcal{R}$.
In the neighborhood of $\partial \mathcal{R}$, $\bar{V}$ is lower (i.e. better) in $\mathcal{R}_1$ than in $\mathcal{R}_2$.

After training the critic with the first batch of TO episodes, the network smooths out the discontinuity (see Fig.~\ref{fig:f1}) in a region $\mathcal{N}$ around $\partial \mathcal{R}$, where the critic either underestimates or overestimates the true value $\bar{V}$:
\begin{align}
V( \bar{x} | \theta_V) > \bar{V}(\bar{x}) \qquad \forall \bar{x} \in \mathcal{N} \cap \mathcal{R}_1 \\ 
V( \bar{x} | \theta_V) < \bar{V}(\bar{x}) \qquad \forall \bar{x} \in \mathcal{N} \cap \mathcal{R}_2
\end{align}
Therefore in $\mathcal{N}$ the critic's gradient will point towards $\mathcal{R}_2$.
Due to this gradient, during the policy improvement phase the actor can improve in $\mathcal{N} \cap \mathcal{R}_2$, learning to steer the state toward $\mathcal{R}_1$ rather than $\mathcal{R}_2$.
This example reveals that the regions near value function's discontinuities hold great potential for policy improvement.

\subsection{General case: Discontinuous Value Function}
The phenomenon shown in Section~\ref{sec:1Dex} frequently occurs in problems with multiple local minima. Each local minimum defines a basin of attraction where the value function is continuous, while discontinuities typically appear at the boundaries between basins.
These discontinuities highlight regions with great potential for policy improvement. In contrast, sampling initial states far from such discontinuities likely leads the actor to simply imitate TO.

\subsection{Detecting Informative Regions via Critic Uncertainty}
Approximating the value function close to the discontinuities is particularly challenging for the critic network, which cannot accurately represent abrupt changes due to its continuous activation functions. As a result, the critic tends to incur larger errors near these boundaries.
We suggest leveraging the uncertainty estimation of the critic's output to identify the value's discontinuities. An additional neural network, referred to as \emph{std-critic}, is introduced to predict the standard deviation of the critic. Following~\cite{stdlearning}, this network is trained at the end of each actor–critic update phase by minimizing the negative log-likelihood of a normal distribution:
\begin{mini!}|l|[2]<b>
{\scriptstyle{\theta^{\text{std}}}}
{\ln(V^{\text{std}}(\tilde{x}|\theta^{\text{std}})) + \frac{1}{2}\frac{(\bar{V} - V(\tilde{x}|\theta^V))^2}{V^{\text{std}}(\tilde{x}|\theta^{\text{std}})^2}}
{\label{std_critic_update}}{}
\end{mini!}
where $V^{\text{std}}(\cdot)$ and $\theta^{\text{std}}$ are the \emph{std-critic} network and its parameters. 
The first term of this loss pushes $V^{std}$ to be small everywhere, while the second term pushes $V^{\text{std}}$ to increase when the critic's error is large.
We train this network after the critic to avoid issues during the early stages of learning.
The initial states with the highest potential for policy improvement are then selected according to their predicted uncertainty $V^{\text{std}}(\tilde{x})$.

\subsection{Algorithm Overview}
\begin{figure*}[t]
\centering
\includegraphics[width = 0.9\textwidth]{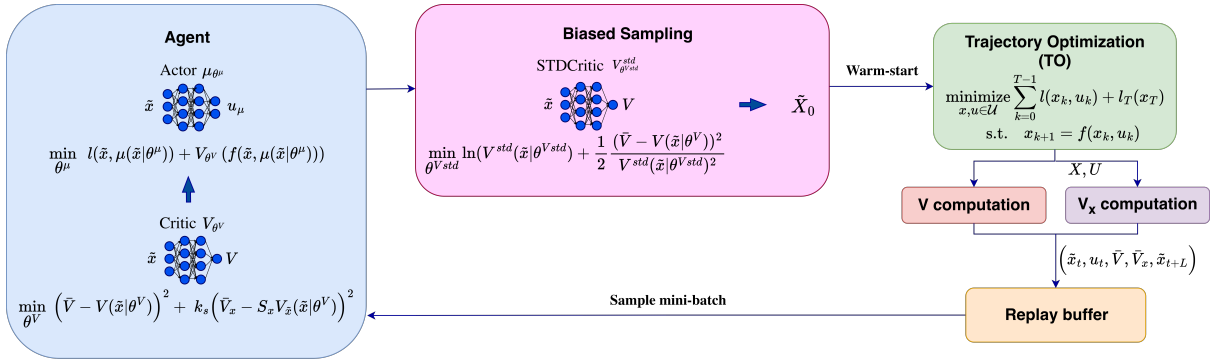}
\caption{Overview of CACTO-BIC with biased initial-state sampling.}
\label{fig:CACTO_scheme}
\end{figure*}

An overview of the CACTO-BIC (Biased Initial Conditions) algorithm is illustrated in Fig.~\ref{fig:CACTO_scheme}.
During the first iteration, as in the original CACTO framework~\cite{CACTO}, initial states for the TO problems are sampled uniformly at random. From the second iteration onward, a set of $10N$ candidate initial states is sampled and ranked based on $V^{\text{std}}$. The top $N$ samples are then selected for the next TO batch. Moreover, since trajectories starting from these initial states provide more information, we can reduce the number of problems solved from the second iteration (see Section~\ref{ssec:bics_results} for details).

\section{GPU-based computation}
To reduce the computation time of the algorithm by leveraging GPUs, the entire framework was migrated to JAX, a Python library for accelerator-oriented array computation and program transformation~\cite{jax2018github}. 
The migration is beneficial in two ways: first, migrating to GPU the neural network training---accounting for about 90\% of CACTO’s total computation time--- yields significant speedups; second, performing TO directly on GPU eliminates any CPU-GPU data-transfer overhead, in addition to reducing TO's computation time, which significantly grows with the complexity of the system.

Migrating to GPU-based computation required some adaptations to meet the constraints of parallel processing: GPUs require fixed-size arrays and uniform computation across batches to enable efficient vectorization, leading to the following two key challenges.

%
\subsection{Fixed number of iterations} 
\label{ssec:iteration_number}
Because GPUs execute batched computations in parallel, all TO problems in a batch must undergo the same number of optimization iterations. Consequently, the maximum number of iterations, $max\_iter$, becomes a crucial hyperparameter. Setting this value too low may lead to premature termination, and hence to collect too few or insufficiently informative samples. In contrast, setting it too large can cause longer runtimes, since hard-to-converge problems dominate the batch’s overall computation time, slowing the entire pipeline. Moreover, the ideal number of iterations may vary as the policy improves, as better initial guesses generally lead to faster convergence.

To determine the maximum number of iterations, we suggest to solve a large set of problems using a naive warm-start and a high iteration limit ($max\_iter=1000$), generating a dataset of the iterations needed to converge. The maximum number of iterations can then be set to the 99th percentile of iteration counts for the first iteration of CACTO (where the naive warm-start is used) and to the 50th percentile for subsequent iterations (where CACTO’s actor provides the warm-start). 

\subsection{Regularization for matrix inversion}
In iLQR~\cite{jacobson1970differential}, it is crucial to regularize the Hessian of the value function so that it is positive definite. 

The standard regularization
consists in increasing the local control-cost Hessian with a diagonal term:
$
    \tilde{Q}_{uu} = Q + \mu I_{m},
$
where $\mu$ plays the role of a Levenberg-Marquardt parameter.
During the backward pass, if the line search fails, 
$\mu$ is increased and the backward pass is retried, otherwise, $\mu$ is decreased.

This approach is computationally efficient when solving problems sequentially on a CPU. However, it becomes impractical when solving multiple problems in parallel on a GPU, where  such iterative tuning can significantly slow down the batched computation.

This issue is addressed through regularization based on the eigenvalue decomposition. The eigenvalues $S$ and corresponding eigenvectors $W$ of the state-cost Hessian and the control-cost Hessian are computed, the eigenvalues are clipped from below by a user-defined constant $\epsilon>0$, $S^\prime = \max(S,\epsilon)$, and the matrices are reconstructed: 
\begin{align}
  Q_+ &= W \, \text{diag}(S^\prime) \, W^T, \quad
  Q_{psd} &= \frac{1}{2} (Q_+ + Q_+^T)
\end{align}
This procedure handles poorly-conditioned Hessians, while avoiding additional iterative steps. As a result, it is better suited for large-batch processing on GPUs.

\subsection{Implementation Details}
The system dynamics and cost functions, implemented using the CasADi library~\cite{casadi} in CACTO-SL~\cite{alboni2024cacto}, were converted into JAX-compatible functions through the Jaxadi library~\cite{jaxadi2024}. For TO, we replaced the previous solver with an iLQR implementation from the Trajax library~\cite{trajax}, which provides the additional benefit of computing the gradient of the value function (used for training the critic network) while solving the optimal control problem. The neural networks were implemented using the Flax library~\cite{flax2020github}, allowing the entire pipeline to remain on the GPU.
Our new implementation is open-source and available on the \href{https://anonymous.4open.science/r/cacto-487E/README.md}{GitHub page of the project}. 

\section{RESULTS}
This section presents our evaluation of CACTO-BIC.
First, we assess the impact of the biased initial-state sampling on data efficiency (Section~\ref{ssec:bics_results}). 
Second, we analyze the computational benefits of the GPU-based implementation (Section~\ref{ssec:gpu_results}). 
Third, we compare CACTO-BIC with a state-of-the-art RL algorithm (Section~\ref{ssec:ppo_results}). 
Finally, we demonstrate the scalability of the approach through experiments on a high-dimensional quadruped robot, AlienGO~\cite{aliengo} (Section~\ref{ssec:aliengo_results}).

\subsection{Biased Initial State Sampling} 
\label{ssec:bics_results}
\begin{figure}[tbp]
     \makebox[\columnwidth][c]{\includegraphics[width = 0.75\columnwidth]{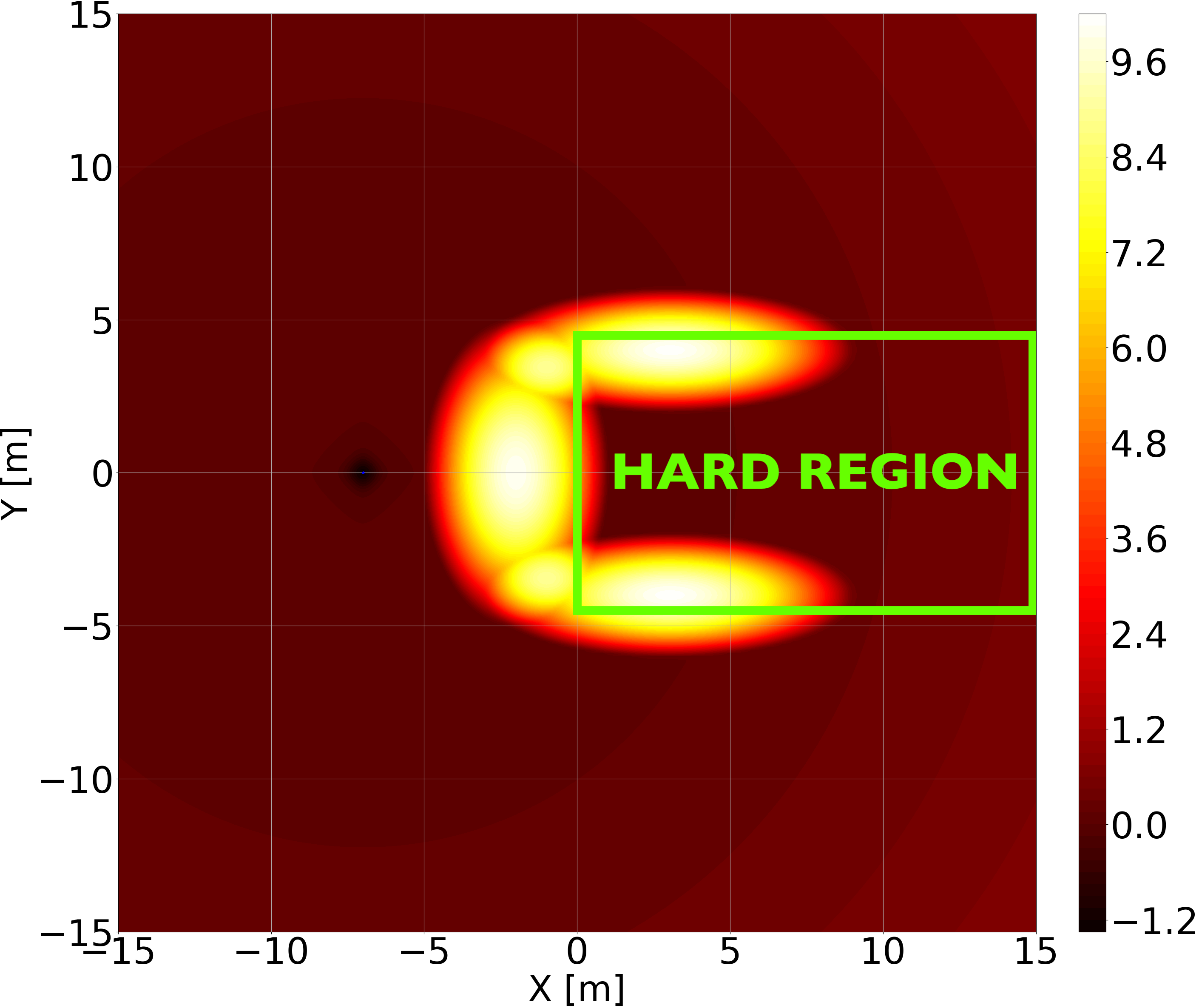}} 
    \caption{Cost function excluding the control effort term, with target set at $[-7,0]$.}
    \label{fig:CostFunction_comp}
\end{figure}

We evaluate the proposed exploration method on the same benchmark scenarios used in~\cite{CACTO, alboni2024cacto}.
The task consists in minimizing the distance between the robot’s end-effector and a target, while avoiding three elliptical obstacles (encoded with large penalties) and minimizing control effort. An additional reward is provided in the neighborhood of the target, as shown in Fig.~\ref{fig:CostFunction_comp}.
When the system starts from the \emph{Hard Region} (highlighted in Fig.~\ref{fig:CostFunction_comp}), it becomes challenging for a gradient-based solver to converge to the globally optimal solution.

We consider three systems of increasing complexity: a point mass with state $(x, y, v_x, v_y, t) \in \mathbb{R}^5$ and control $(a_x, a_y) \in \mathbb{R}^2$, a jerk-controlled version of the \emph{Dubins car model} \cite{dubins} with state $(x,y,\theta,v,a,t) \in \mathbb{R}^6$ and control $(\omega,j) \in \mathbb{R}^2$, and a 3-degree-of-freedom (DoF) planar manipulator with a 7-dimensional state space and a 3-dimensional control input.
We compare three algorithms:
\begin{enumerate}
    \item CACTO as presented in~\cite{alboni2024cacto};
    \item CACTO-BIC: CACTO with biased initial-state sampling and a reduced number of TO episodes (25\%) from the 2nd iteration onward;
    \item CACTO with reduced TO episodes (as CACTO-BIC), but without biased initial-state sampling.
\end{enumerate}

\begin{figure}[tb]
     \centering
     {\input{Fig/BICS_DI}}\par
     \vspace{0.25cm}
     {\input{Fig/BICS_C}}\par
     \vspace{0.25cm}
     {\input{Fig/BICS_M_HR}}
    \caption{Median (across 5 runs) of the mean cost (across initial conditions) starting from the Hard Region for the point mass (top), the \emph{Dubins car}, and the manipulator (bottom). Shaded areas represent first and third quartiles. Data are sampled every \textit{n} updates. When multiple measurements occur at the same TO episode count, the last one is plotted.}
    \label{fig:CACTO-SL-BICScomparison}
\end{figure}
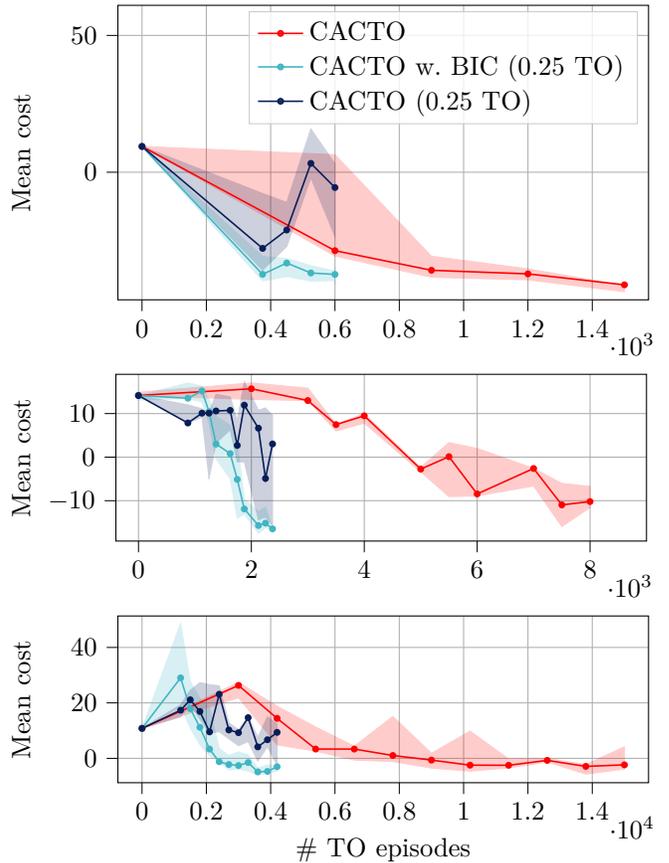

All algorithms run for the same number of learning iterations (i.e. updates of the neural networks). Fig.~\ref{fig:CACTO-SL-BICScomparison} shows the median (across 5 runs) of the mean cost (across initial conditions) as a function of the number of TO episodes. The results show that CACTO-BIC achieves comparable performance using $30-40\%$ of the number of TO episodes. 
However, this improvement comes with increased training time due to the additional std-critic network. With CACTO-BIC, the percentage of computation time devoted to training the networks rises to $93-94\%$.
This motivates moving to the GPU.

\subsection{Computational Efficiency of GPU Computation}
\label{ssec:gpu_results}
First, we assessed the speedup achievable when solving a batch of TO problems on a GPU for different batch sizes: $[10, 50, 100, 250, 500, 1000, 5000, 10000]$. 

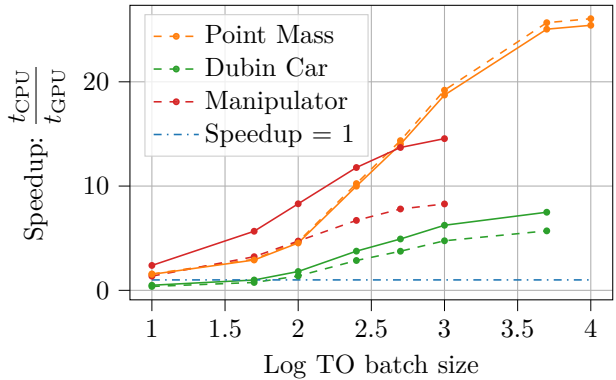
\begin{figure}[tb]
     \makebox[\columnwidth][c]
     {\input{Fig/TOBatchSizeSpeedup}}
    \caption{Speedups for TO on GPU (w.r.t. CPU) for the point mass, the \emph{Dubins car}, and the manipulator. Dashed lines represent speedups with random initial states and maximum number of iterations, selected as described in Section~\ref{ssec:iteration_number}, while solid lines represent speedups with initial conditions selected using CACTO-BIC and maximum number of iterations set as in the two versions.}
    \label{fig:TOBatchSize}
\end{figure}

Fig.~\ref{fig:TOBatchSize} shows that while the CPU computation time scales approximately linearly with batch size, the GPU benefits from parallelization: the time per problem decreases as the batch size increases, eventually saturating, demonstrating the advantage of GPU-based TO. 

We now analyze the speedup achieved through the new GPU-accelerated implementation of CACTO-BIC. The number of TO problems solved in parallel was 300, 500, and 550 in the first iteration for the point mass, Dubins car and manipulator, respectively. From the second iteration onward, the batch size was reduced to 25\%. The initial time is set to 0 in each TO instance.
Table~\ref{tab:comp_comp_time} reports the computation times to perform the same updates using three versions of CACTO-BIC: a single-thread CPU version, a multi-thread CPU version, and the novel GPU version.
\begin{table}[tbp]
\caption{Comparison between CPU and GPU versions of CACTO-BIC. GPU runtimes do not include the warm-up phase required for the JIT compilation of the TO solver.}
\label{tab:comp_comp_time}
\centering
\begin{tabular}{|c|c|c|c|}
\hline
\textbf{System} (nb. of updates) & \multicolumn{2}{c|}{\textbf{CPU}} & \textbf{GPU} \\
\cline{2-3}
& 1 core & 10 cores & \\
\hline
\makecell{\textbf{Point Mass} (30k)}
& 23 min & 22 min & 44 s \\ 
\hline
\makecell{\textbf{Dubins Car} (200k)}
& 3 h 44 min & 3 h 34 min & 7 min \\ 
\hline
\makecell{\textbf{Manipulator} (330k)}
& 5 h 35 min & 5 h 15 min & 6 min \\ 
\hline
\end{tabular}
\end{table}
This test was performed on a workstation equipped with an AMD Ryzen 9 7950X CPU, 192 GB of RAM, and an NVIDIA RTX 6000 GPU with 48 GB VRAM, running on Ubuntu 22.04. 

The results show that using 10 cores to parallelize the TO problems has little effect on the total computation time as most time is spent for training the neural networks. The novel GPU version of CACTO-BIC achieves instead remarkable speedups ($\approx$30x for the point mass and Dubins car, and $\approx$56x for the manipulator). 

Although GPUs have far more cores than CPUs, the resulting speedups are smaller than this hardware difference suggests. Several factors contribute to this limitation. First, the TO problems are solved in batches on the GPU. Therefore a few hard instances may dominate the total computation time and cap the achievable speedup. Second, computations rely on reduced numerical precision. Although this improves raw throughput, it can introduce numerical instability and increase the number of required iterations.

By analyzing the time spent in the TO phase ({creating the warm-start} and {solving TO problems}) and in the RL phase ({training networks}), we observe that their relative contribution varies across systems, but remains comparatively balanced. Specifically, the TO and RL phases account for approximately 42\% and 53\% of the total time in the point mass, 61\% and 37\% in the Dubins car, and 22\% and 75\% in the manipulator.
It follows that the speedup comes mainly from training the networks on the GPU, which is 51-85$\times$ faster than on the CPU in our tests.
In contrast, the speedup in the TO phase is modest (3-5$\times$) for the Dubins car and point mass, but it becomes significant (19$\times$) for the most complex system.

\subsection{Comparison with PPO}
\label{ssec:ppo_results}
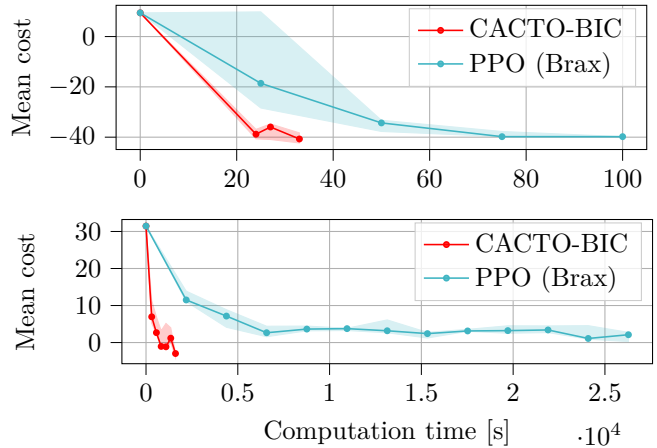
\begin{figure}[tb]
     \makebox[\columnwidth][c]
     {\input{Fig/CvsB_DI.tex}}\par
     \vspace{0.25cm}
     { \input{Fig/CvsB_M_HR.tex}} 
    \caption{Warm-start provider comparison: Median (across 5 runs) of the mean cost (across initial conditions) starting from the Hard Region for the point mass (top), and the manipulator (bottom). Shaded areas represent first and third quartiles.}
    \label{fig:comparison_WS}
\end{figure}

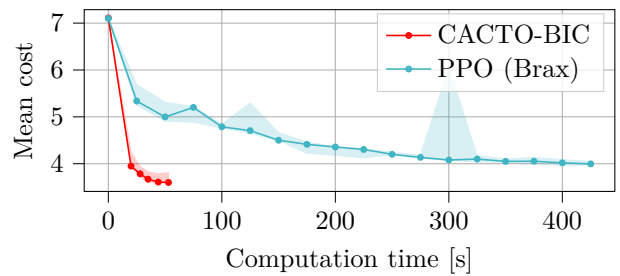
\begin{figure}[tb]
     \makebox[\columnwidth][c]
     {\input{Fig/CvsB_R.tex}}
    \caption{Policy comparison: Median (across 5 runs) of the mean cost (across initial conditions) for a customized \textit{Reacher} environment. Shaded areas represent first and third quartiles.}
    \label{fig:comparison_policy}
\end{figure}

To evaluate CACTO-BIC w.r.t. state-of-the-art RL, we selected Proximal Policy Optimization (PPO) \cite{PPO} as a benchmark. For a fair comparison, we employed the fully GPU-based implementation of PPO provided by BRAX \cite{brax2021github}. Our analysis focuses on two aspects: first, learning a warm-start policy for a problem that exhibits local minima; second, learning a control policy for a classic benchmark problem, a customized version of the \emph{Reacher} environment. Fig.~\ref{fig:comparison_WS} reports the mean cost obtained using CACTO-BIC's and PPO's policies as warm-start for TO in the Point Mass and Manipulator environments. CACTO-BIC converged in one-third of the time in the first environment and in just 7\% in the second one. 
In Fig.~\ref{fig:comparison_policy} we compare CACTO-BIC and PPO in the Reacher Environment. Also in this case, CACTO-BIC achieves a similar cost in $\approx$10\% of the computation time.

\subsection{Hardware experiments}
\label{ssec:aliengo_results}
To evaluate CACTO's scalability and its potential for real-time applications we tested it on AlienGO, a quadrupedal robot featuring  12 degrees of freedom~\cite{aliengo}.
We address the problem of navigation in confined environments, with even terrain, and the presence of a moving obstacle. The robot must reach a moving target (see Fig.~\ref{fig:setup}) while avoiding collisions with both the moving obstacle (a sphere of radius 0.5 m) and the walls of the room (a rectangle with sides ranging from 2 m to 10 m). We consider a fixed trotting gait with alternating diagonal leg pairs making contact with the ground.

Since legged robots have non-differentiable dynamics~\cite{wensing2023optimization}, which are not compatible with gradient-based TO, we adopt a hierarchical approach. We combine a high-level policy relying on a simplified differentiable model, with a low-level policy using the complete robot dynamics. 
The high-level policy is trained with CACTO-BIC, while the low-level policy is trained with the RL algorithm CAT~\cite{chane2024cat}, which can handle non-differentiable dynamics. In practice, CACTO-BIC's actor is used as a standalone policy to provide reference trajectories to the low-level policy. 

\subsubsection{High-level policy}
To train this policy, we employed a nonlinear version of the Linear Inverted Pendulum Model~\cite{kajita2003biped}. 
The state $x \in \mathbb{R}^{15}$ comprises the 2D offsets of the front and rear support feet relative to the associated shoulders $\Delta p_f$ and $\Delta p_r$, the Center of Mass (CoM) position $c \in \mathbb{R}^2$ and velocity $\dot{c} \in \mathbb{R}^2$ on the horizontal plane, the step index $s_{idx} \in \mathbb{N}$, which encodes both the current contact phase and time, the obstacle position $c_{obs} \in \mathbb{R}^2$ and the location of the four walls, expressed as offsets with respect to the global reference frame $\Delta_{walls} \in \mathbb{R}^4$.
\begin{equation}
x \triangleq (\Delta p_f, \Delta p_r, c, \dot{c}, s_{idx}, c_{obs}, \Delta_{walls}) \in \mathbb{R}^{15}.
\end{equation}
The target position is assumed to be the origin.

Assuming a trotting gait and constant Center of Pressure (CoP) during each contact phase, the control vector $u \in \mathbb{R}^6$ includes the offsets of  front and rear support feet w.r.t. the associated shoulder in the next contact phase, $\Delta p_f ^ +, \Delta p_r ^ + \in \mathbb{R}^2$, a scalar $\alpha \in [0,1]$ that expresses the CoP as a convex combination of the support foot positions, and the contact phase duration $\delta t$:
\begin{equation}
u \triangleq (\Delta p_f^+, \Delta p_r^+, \alpha, \delta t) \in \mathbb{R}^{6}
\end{equation}

The cost function penalizes the CoM-target distance, the CoM velocities, and a barrier-like cost penalizes CoM velocities beyond prescribed bounds. A smooth logarithmic reward is used to encourage reaching a narrow region around the target. Control regularization discourages deviations from nominal values: zero foot displacements, centered CoP ($\alpha=0.5$), and nominal contact-phase duration ($\delta t=0.375$ s). Obstacle avoidance is enforced through smooth logarithmic penalties. The algorithm took $\approx$109 s to converge. In this scenario, leveraging the GPU for solving the TO problems yields a substantial speedup, roughly 345$\times$. 

\subsubsection{Low-level policy}
\label{lowlevel}

The low-level policy is trained using a constrained RL algorithm \cite{chane2024cat}, which builds upon PPO \cite{PPO}, employing Isaac Gym for GPU simulation. The goal of this policy is to track the references generated by the high-level policy. During training, the high-level policy is rolled out in open loop, and updated only at the beginning of each new contact phase to enhance stability and robustness~\cite{villa2017model}. 

The low-level policy receives as observations: the base state (position and orientation, linear and angular velocity, and projected gravity), the target base position and linear velocity, yaw orientation error, foot placement errors, a binary flag indicating the active diagonal contact pair, the remaining time in the current gait phase, joint positions, velocities and the previous action.

The reward function encourages accurate tracking of base position and orientation, linear and angular velocity, and the desired foot contact locations. Regularization terms are included to improve smoothness. 
All safety and style-related constraints are divided into soft constraints (which define a termination probability as a function of the constraint violation) and hard constraints (which cause immediate termination of the episode).



\subsubsection{Aliengo Simulation and Hardware results}
\begin{figure}[tbp]
     \makebox[\columnwidth][c]{\includegraphics[width = 0.9\columnwidth]{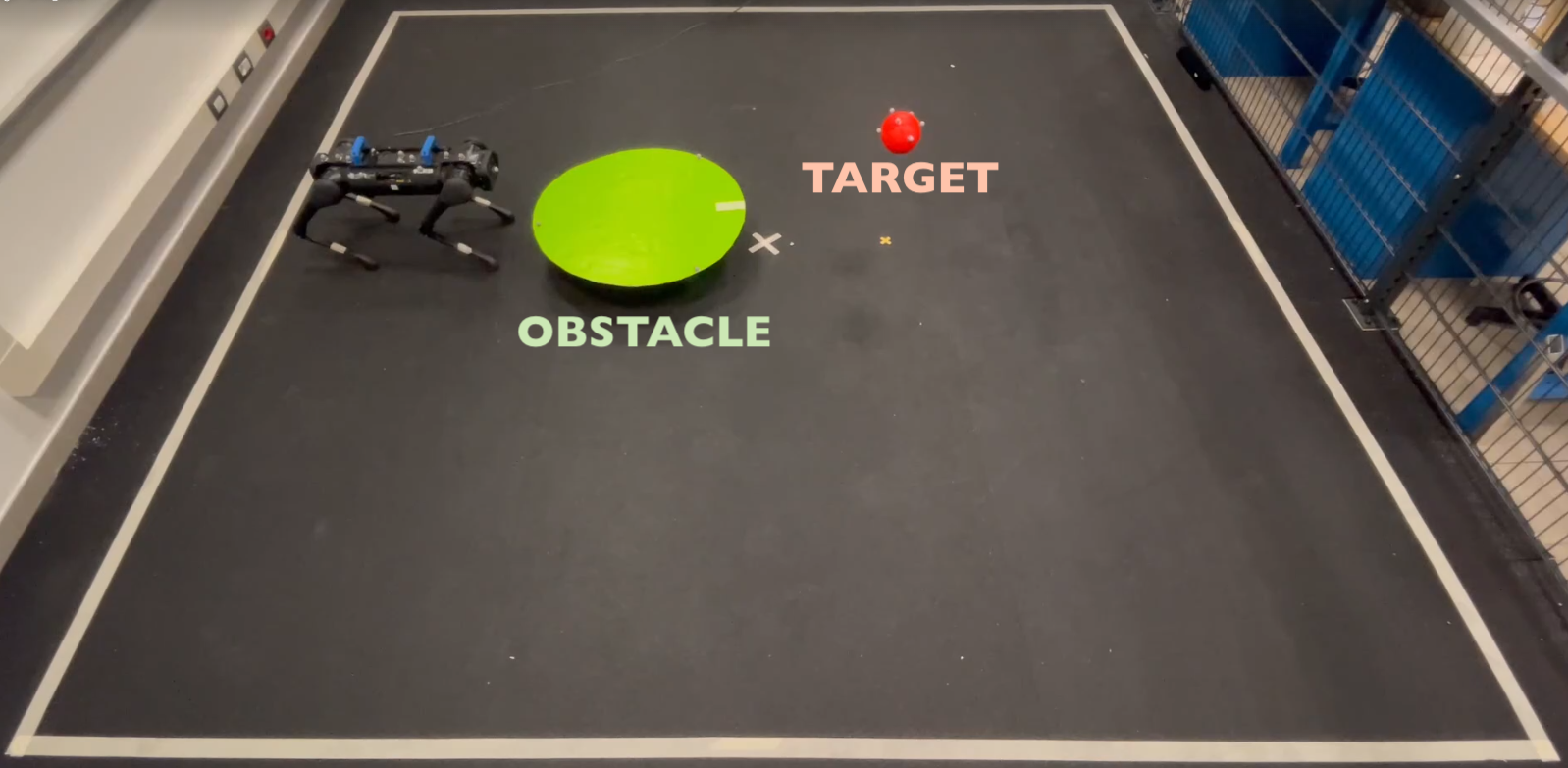}} 
    \caption{Experimental setup: The lines define a 4 m $\times$ 4 m operational area.}
    \label{fig:setup}
\end{figure}

The experiments are conducted in an indoor area of 4 m$\times$4 m (see Fig.~\ref{fig:setup}). The obstacle and the target are either stationary or manually actuated, depending on the experiment. 
State feedback is obtained from a motion capture system, which measures  position and orientation of the robot, the obstacle and the target. Velocities are estimated by finite differencing and low-pass filtering. 

Results are shown in the accompanying video.

\section{CONCLUSIONS}
This paper presented CACTO-BIC, an extension of CACTO designed to improve scalability and computational efficiency in combined TO-RL framework by biasing the initial-state sampling using value function properties and leveraging GPU acceleration.

Experimental results demonstrate that the proposed sampling strategy effectively identifies state-space regions where actor policy improvement is more likely, leading to a $2.5-3.5\times$ increase in sample efficiency. In addition, GPU-based computation achieves 30-250$\times$ speedup compared to CACTO, with increasing benefits as system complexity grows. 
When compared to PPO, CACTO-BIC achieves similar final costs requiring only $7-30\%$ of the training time. 
Finally, experiments on the AlienGO quadruped robot show that CACTO-BIC scales to high-dimensional robotic systems and can be effectively used for real robot control.

Future works will focus on extending the proposed approach to handle constraints using augmented Lagrangian formulations \cite{crl}. In addition, we will explore the integration of sampling-based optimization techniques, such as MPPI \cite{mppi}, to tackle non-differentiable dynamics. Finally, applying domain randomization may improve robustness to uncertain dynamics enhancing generalization to real-world systems \cite{domrand}.

\bibliographystyle{IEEEtran}
\bibliography{references}

\end{document}

%% file: Fig/BICS_DI.tex
\begin{tikzpicture}

\definecolor{darkgray176}{RGB}{176,176,176}
\definecolor{lightgray204}{RGB}{204,204,204}
\definecolor{mediumturquoise64181195}{RGB}{64,181,195}
\definecolor{midnightblue82988}{RGB}{8,29,88}

\begin{axis}[
width=1\linewidth,
height=5.5cm,
legend cell align={left},
legend style={fill opacity=0.8, draw opacity=1, text opacity=1, draw=lightgray204, row sep=0pt},
tick align=outside,
tick pos=left,
x grid style={darkgray176},
xmajorgrids,
xmin=-75, xmax=1575,
xtick style={color=black},
scaled x ticks = {base 10:-3},
y grid style={darkgray176},
ylabel={Mean cost},
ymajorgrids,
ymin=-46.6360807767274, ymax=61,
extra y ticks={-100},
ytick style={color=black},
x tick scale label style={
    at={(0.925,-0.225)},
    anchor=south west,
    yshift=4pt}
]
\path [draw=red, fill=red, opacity=0.2]
(axis cs:0,9.42711082140595)
--(axis cs:0,9.42711082140595)
--(axis cs:600,-30.6398475389518)
--(axis cs:900,-38.3870572703966)
--(axis cs:1200,-39.343681240459)
--(axis cs:1500,-43.6626359622984)
--(axis cs:1500,-42.0033268099325)
--(axis cs:1500,-40.0198098006524)
--(axis cs:1500,-41.4209791741542)
--(axis cs:1200,-35.3154565577383)
--(axis cs:900,-30.6957660540385)
--(axis cs:600,6.41207801976498)
--(axis cs:0,9.42711082140595)
--cycle;

\path [draw=mediumturquoise64181195, fill=mediumturquoise64181195, opacity=0.2]
(axis cs:0,9.4271108212905)
--(axis cs:0,9.4271108212905)
--(axis cs:375,-39.7168850255748)
--(axis cs:450,-38.1259335098959)
--(axis cs:525,-39.7964538450579)
--(axis cs:525,-39.7950540699404)
--(axis cs:600,-39.5718134384873)
--(axis cs:600,-40.9235263641537)
--(axis cs:600,-34.7340942217205)
--(axis cs:600,-34.7340942217205)
--(axis cs:600,-36.1128269806174)
--(axis cs:525,-34.1121602757411)
--(axis cs:450,-31.534101196855)
--(axis cs:375,-30.7803437905337)
--(axis cs:0,9.4271108212905)
--cycle;

\path [draw=midnightblue82988, fill=midnightblue82988, opacity=0.2]
(axis cs:0,9.4271108212905)
--(axis cs:0,9.4271108212905)
--(axis cs:375,-35.7835691882398)
--(axis cs:450,-27.1555496909907)
--(axis cs:525,-2.00578765470382)
--(axis cs:600,-23.4490192693016)
--(axis cs:600,-12.8721243382845)
--(axis cs:600,-5.19197066947129)
--(axis cs:600,-5.19197066947129)
--(axis cs:600,3.09115686438541)
--(axis cs:525,15.80626032628)
--(axis cs:450,-11.0395819018545)
--(axis cs:375,-7.77340030733507)
--(axis cs:0,9.4271108212905)
--cycle;

\addplot [semithick, red, mark=*, mark size=1, mark options={solid}]
table {%
0 9.42711082140595
600 -28.6771863650541
900 -35.8393133183918
1200 -37.1582202684398
1500 -41.1526044597217
};
\addlegendentry{CACTO}
\addplot [semithick, mediumturquoise64181195, mark=*, mark size=1, mark options={solid}]
table {%
0 9.4271108212905
375 -37.3185901117168
450 -33.189884820169
525 -36.8219502576598
600 -37.344916908868
};
\addlegendentry{CACTO w. BIC (0.25 TO)}
\addplot [semithick, midnightblue82988, mark=*, mark size=1, mark options={solid}]
table {%
0 9.4271108212905
375 -27.8275417992375
450 -21.1200289550542
525 3.2219233210601
600 -5.5986757107251
};
\addlegendentry{CACTO (0.25 TO)}
\end{axis}

\end{tikzpicture}

%% file: Fig/BICS_C.tex
\begin{tikzpicture}

\definecolor{darkgray176}{RGB}{176,176,176}
\definecolor{lightgray204}{RGB}{204,204,204}
\definecolor{mediumturquoise64181195}{RGB}{64,181,195}
\definecolor{midnightblue82988}{RGB}{8,29,88}

\begin{axis}[
width=1\linewidth,
height=3.8cm,
legend cell align={left},
legend style={fill opacity=0.8, draw opacity=1, text opacity=1, draw=lightgray204, row sep=0pt},
tick align=outside,
tick pos=left,
x grid style={darkgray176},
xmajorgrids,
xmin=-400, xmax=9000, 
xtick style={color=black},
scaled x ticks = {base 10:-3},
y grid style={darkgray176},
ylabel={Mean cost},
ymajorgrids,
ymin=-19.2657146766768, ymax=19.0156237415285,
ytick style={color=black},
x tick scale label style={
    at={(0.925,-0.4)},
    anchor=south west,
    yshift=4pt
}
]
\path [draw=red, fill=red, opacity=0.2]
(axis cs:0,14.89975473466)
--(axis cs:0,14.1332257036077)
--(axis cs:2000,13.2613205136697)
--(axis cs:3000,12.9761904829767)
--(axis cs:3500,5.99645876984661)
--(axis cs:4000,7.796758427066)
--(axis cs:5000,-3.37175063573384)
--(axis cs:5500,-9.05849182285904)
--(axis cs:6000,-8.91010273893671)
--(axis cs:7000,-6.62328468178424)
--(axis cs:7500,-15.9372098907875)
--(axis cs:8000,-11.5711071802413)
--(axis cs:8000,-6.71414012284177)
--(axis cs:8000,-6.71414012284177)
--(axis cs:7500,-6.02981170885749)
--(axis cs:7000,-2.57935562689088)
--(axis cs:6000,1.99458879344934)
--(axis cs:5500,3.32819094644904)
--(axis cs:5000,-2.33875150199564)
--(axis cs:4000,9.95789361281389)
--(axis cs:3500,7.56261906985632)
--(axis cs:3000,15.8310677404047)
--(axis cs:2000,17.0364603035448)
--(axis cs:0,14.89975473466)
--cycle;

\path [draw=mediumturquoise64181195, fill=mediumturquoise64181195, opacity=0.2]
(axis cs:0,14.1332257036077)
--(axis cs:0,13.0964996066486)
--(axis cs:875,13.4710620142817)
--(axis cs:1125,12.6354100237448)
--(axis cs:1250,7.80574670418417)
--(axis cs:1375,-0.406593486789507)
--(axis cs:1625,-4.97826344231376)
--(axis cs:1750,-14.0567772012801)
--(axis cs:1875,-13.036759465809)
--(axis cs:2125,-17.5256538394857)
--(axis cs:2250,-16.1804316861236)
--(axis cs:2375,-16.831748852439)
--(axis cs:2375,-16.3682415097177)
--(axis cs:2375,-16.3682415097177)
--(axis cs:2250,-11.502857926272)
--(axis cs:2125,-12.4954628105324)
--(axis cs:1875,-9.65717658821941)
--(axis cs:1750,3.19722011364524)
--(axis cs:1625,7.32602764617472)
--(axis cs:1375,9.35270024483941)
--(axis cs:1250,13.2629372357992)
--(axis cs:1125,16.2917879593333)
--(axis cs:875,17.0421046971072)
--(axis cs:0,14.1332257036077)
--cycle;

\path [draw=midnightblue82988, fill=midnightblue82988, opacity=0.2]
(axis cs:0,14.1332257036077)
--(axis cs:0,14.1332257036077)
--(axis cs:875,7.83379355895619)
--(axis cs:1125,9.15771520046867)
--(axis cs:1250,-4.81245252504678)
--(axis cs:1375,3.24655759191232)
--(axis cs:1625,6.28927065044845)
--(axis cs:1750,-1.06643128338617)
--(axis cs:1875,-1.10205950903285)
--(axis cs:2125,-14.3677535668834)
--(axis cs:2250,-13.7592965034821)
--(axis cs:2375,-15.5703902911283)
--(axis cs:2375,9.58738466586264)
--(axis cs:2375,9.58738466586264)
--(axis cs:2250,11.2740899743294)
--(axis cs:2125,10.7533318465594)
--(axis cs:1875,17.2755629043373)
--(axis cs:1750,8.47430079337581)
--(axis cs:1625,13.8168384455835)
--(axis cs:1375,14.4306839010813)
--(axis cs:1250,10.2237088383797)
--(axis cs:1125,11.0847611669814)
--(axis cs:875,12.0126783026341)
--(axis cs:0,14.1332257036077)
--cycle;

\addplot [semithick, red, mark=*, mark size=1, mark options={solid}]
table {%
0 14.1629542732551
2000 15.6931110062191
3000 12.9973524509577
3500 7.47292319714952
4000 9.48610766405917
5000 -2.75121613619893
5500 0.106850960907421
6000 -8.44928203283456
7000 -2.60501417709044
7500 -10.9640808779058
8000 -10.2041711297403
};
\addplot [semithick, mediumturquoise64181195, mark=*, mark size=1, mark options={solid}]
table {%
0 14.1332257036077
875 13.5226898673687
1125 15.2007846204137
1250 10.4376341826618
1375 3.00372550352931
1625 0.812963176152132
1750 -5.11898602460446
1875 -11.8977319618623
2125 -15.6968529191522
2250 -15.1558406774643
2375 -16.4704193468681
};
\addplot [semithick, midnightblue82988, mark=*, mark size=1, mark options={solid}]
table {%
0 14.1332257036077
875 7.83649778751844
1125 10.0938385875974
1250 10.1054962089026
1375 10.5943606993
1625 10.7385863566119
1750 2.69697587243428
1875 11.9289035196693
2125 6.64195958299485
2250 -4.89323898157967
2375 3.02448499992247
};
\end{axis}

\end{tikzpicture}

%% file: Fig/BICS_M_HR.tex
\begin{tikzpicture}

\definecolor{darkgray176}{RGB}{176,176,176}
\definecolor{lightgray204}{RGB}{204,204,204}
\definecolor{mediumturquoise64181195}{RGB}{64,181,195}
\definecolor{midnightblue82988}{RGB}{8,29,88}

\begin{axis}[
width=1\linewidth,
height=3.8cm,
legend cell align={left},
legend style={fill opacity=0.8, draw opacity=1, text opacity=1, draw=lightgray204},
tick align=outside,
tick pos=left,
x grid style={darkgray176},
xlabel={\# TO episodes},
xmajorgrids,
xmin=-750, xmax=15750,
xtick style={color=black},
y grid style={darkgray176},
ylabel={Mean cost},
ymajorgrids,
ymin=-8.7880263479271, ymax=51.3470910468905,
ytick style={color=black},
extra y ticks={-20},
extra y tick labels={},
x tick scale label style={
    at={(0.925,-0.4)},
    anchor=south west,
    yshift=4pt
}
]
\path [draw=red, fill=red, opacity=0.2]
(axis cs:0,10.8094265169446)
--(axis cs:0,10.8094265169446)
--(axis cs:3000,21.8827328737924)
--(axis cs:4200,4.85801089142459)
--(axis cs:5400,2.59605770180897)
--(axis cs:6600,-0.602573755219995)
--(axis cs:7800,-1.16619232108437)
--(axis cs:9000,-3.51565803378421)
--(axis cs:10200,-4.5996873266006)
--(axis cs:11400,-3.475757753628)
--(axis cs:12600,-1.25803799061686)
--(axis cs:13800,-5.6865377186767)
--(axis cs:15000,-3.87323622088513)
--(axis cs:15000,4.16125366901069)
--(axis cs:15000,4.16125366901069)
--(axis cs:13800,-2.24439620545116)
--(axis cs:12600,-0.0372830818263338)
--(axis cs:11400,-0.215388330723566)
--(axis cs:10200,9.81255497903252)
--(axis cs:9000,1.8450311904573)
--(axis cs:7800,15.101254190625)
--(axis cs:6600,4.38501737877048)
--(axis cs:5400,11.294166981108)
--(axis cs:4200,18.8166590439408)
--(axis cs:3000,27.4052854913349)
--(axis cs:0,10.8094265169446)
--cycle;

\path [draw=mediumturquoise64181195, fill=mediumturquoise64181195, opacity=0.2]
(axis cs:0,10.8094265169446)
--(axis cs:0,10.8094265169446)
--(axis cs:1200,18.7867482732937)
--(axis cs:1500,10.8668808783697)
--(axis cs:1800,4.87718408882579)
--(axis cs:2100,0.997277013870994)
--(axis cs:2400,-1.26422160188577)
--(axis cs:2700,-4.20231546480903)
--(axis cs:3000,-3.88454189317451)
--(axis cs:3300,-4.60053278063588)
--(axis cs:3600,-6.05461192088994)
--(axis cs:3900,-5.29768146117613)
--(axis cs:4200,-4.6323550920416)
--(axis cs:4200,-2.86913584564339)
--(axis cs:4200,-2.86913584564339)
--(axis cs:3900,-2.34677474464862)
--(axis cs:3600,-3.49673657345937)
--(axis cs:3300,1.24675647840644)
--(axis cs:3000,2.33195141174185)
--(axis cs:2700,1.02287641369276)
--(axis cs:2400,3.57136272708514)
--(axis cs:2100,18.2241518252163)
--(axis cs:1800,21.537035341651)
--(axis cs:1500,27.7140261208898)
--(axis cs:1200,48.6136766198534)
--(axis cs:0,10.8094265169446)
--cycle;

\path [draw=midnightblue82988, fill=midnightblue82988, opacity=0.2]
(axis cs:0,10.8094265169446)
--(axis cs:0,10.8094265169446)
--(axis cs:1200,14.9171223965445)
--(axis cs:1500,19.4633114857562)
--(axis cs:1800,14.1695558633526)
--(axis cs:2100,9.22072730595836)
--(axis cs:2400,9.57034999847736)
--(axis cs:2700,7.0600274463018)
--(axis cs:3000,4.55271402337857)
--(axis cs:3300,6.05181385200895)
--(axis cs:3600,-0.964847421806414)
--(axis cs:3900,4.15412107895311)
--(axis cs:4200,-2.42789453131447)
--(axis cs:4200,12.4199985186624)
--(axis cs:4200,12.4199985186624)
--(axis cs:3900,14.9597743315907)
--(axis cs:3600,7.53379365063748)
--(axis cs:3300,15.0657938451236)
--(axis cs:3000,10.3347125657665)
--(axis cs:2700,12.7782932524154)
--(axis cs:2400,26.1664734999404)
--(axis cs:2100,26.5832460366441)
--(axis cs:1800,27.3120962379717)
--(axis cs:1500,24.5092282120595)
--(axis cs:1200,19.3320302269201)
--(axis cs:0,10.8094265169446)
--cycle;

\addplot [semithick, red, mark=*, mark size=1, mark options={solid}]
table {%
0 10.8094265169446
3000 26.3069758710849
4200 14.4207158684396
5400 3.34346531989001
6600 3.34044798628948
7800 1.02732106056184
9000 -0.634707609209738
10200 -2.46324925285557
11400 -2.50320696854434
12600 -0.731917253185422
13800 -2.90198673165529
15000 -2.35315827684524
};
\addplot [semithick, mediumturquoise64181195, mark=*, mark size=1, mark options={solid}]
table {%
0 10.8094265169446
1200 29.0287083680299
1500 17.8773099791344
1800 11.1903782712242
2100 3.33920141630799
2400 -1.22141459573364
2700 -2.21968238252886
3000 -2.58159686465266
3300 -1.50961535069631
3600 -4.86876763071541
3900 -4.70531651348126
4200 -3.00806276746335
};
\addplot [semithick, midnightblue82988, mark=*, mark size=1, mark options={solid}]
table {%
0 10.8094265169446
1200 17.3493782999097
1500 21.1120550611935
1800 16.8970278993759
2100 9.53936331226369
2400 23.1201207870161
2700 10.2443242268551
3000 9.25294149771879
3300 14.6772158177542
3600 4.13501776435599
3900 6.71489329857775
4200 9.36895858446096
};
\end{axis}

\end{tikzpicture}

%% file: Fig/TOBatchSizeSpeedup.tex
\begin{tikzpicture}

\definecolor{crimson2143940}{RGB}{214,39,40}
\definecolor{darkgray176}{RGB}{176,176,176}
\definecolor{darkorange25512714}{RGB}{255,127,14}
\definecolor{forestgreen4416044}{RGB}{44,160,44}
\definecolor{lightgray204}{RGB}{204,204,204}
\definecolor{steelblue31119180}{RGB}{31,119,180}

\begin{axis}[
height=5.5cm,
width=8cm,
legend cell align={left},
legend style={
  fill opacity=0.8,
  draw opacity=1,
  text opacity=1,
  at={(0.03,0.97)},
  anchor=north west,
  draw=lightgray204
},
tick align=outside,
tick pos=left,
x grid style={darkgray176},
xlabel={Log TO batch size},
xmajorgrids,
xmin=0.85, xmax=4.15,
xtick style={color=black},
y grid style={darkgray176},
ylabel={Speedup: \(\displaystyle \frac{t_\mathrm{CPU}}{t_\mathrm{GPU}}\)},
ymajorgrids,
ymin=-0.911570126433402, ymax=27.3366411839655,
ytick style={color=black}
]
\addplot [semithick, dashed, darkorange25512714, mark=*, mark size=1, mark options={solid}]
table {%
1 1.57142857142857
1.69897000433602 2.98192771084337
2 4.65334900117509
2.39794000867204 10.2378490175801
2.69897000433602 14.3582306018854
3 19.1953465826466
3.69897000433602 25.6709451575263
4 26.0526315789474
};
\addlegendentry{Point Mass}
\addplot [semithick, dashed, forestgreen4416044, mark=*, mark size=1, mark options={solid}]
table {%
1 0.37243947858473
1.69897000433602 0.754337440281619
2 1.37488542621448
2.39794000867204 2.86150324303701
2.69897000433602 3.74953130858643
3 4.75209884365595
3.69897000433602 5.70342205323194
};
\addlegendentry{Dubin Car}
\addplot [semithick, dashed, crimson2143940, mark=*, mark size=1, mark options={solid}]
table {%
1 1.35955056179775
1.69897000433602 3.22666666666667
2 4.7265625
2.39794000867204 6.70731707317073
2.69897000433602 7.80141843971631
3 8.28483396097227
};
\addlegendentry{Manipulator}
\addplot [semithick, forget plot, darkorange25512714, mark=*, mark size=1, mark options={solid}]
table {%
1 1.53333333333333
1.69897000433602 2.90963855421687
2 4.54054054054054
2.39794000867204 9.98965873836608
2.69897000433602 14.010152284264
3 18.7300048473097
3.69897000433602 25.0486192143135
4 25.4210526315789
};
\addplot [semithick, forget plot, forestgreen4416044, mark=*, mark size=1, mark options={solid}]
table {%
1 0.489137181874612
1.69897000433602 0.990696504903193
2 1.80568285976169
2.39794000867204 3.75810759252194
2.69897000433602 4.92438445194351
3 6.24108981466815
3.69897000433602 7.49049429657795
};
\addplot [semithick, forget plot, crimson2143940, mark=*, mark size=1, mark options={solid}]
table {%
1 2.3876404494382
1.69897000433602 5.66666666666667
2 8.30078125
2.39794000867204 11.7793791574279
2.69897000433602 13.7008381689233
3 14.5498117083191
};
\addplot [semithick, steelblue31119180, dash dot]
table {%
1 1
1.69897000433602 1
2 1
2.39794000867204 1
2.69897000433602 1
3 1
3.69897000433602 1
4 1
};
\addlegendentry{Speedup = 1}
\end{axis}

\end{tikzpicture}

%% file: Fig/CvsB_DI.tex
\begin{tikzpicture}

\definecolor{darkgray176}{RGB}{176,176,176}
\definecolor{lightgray204}{RGB}{204,204,204}
\definecolor{mediumturquoise64181195}{RGB}{64,181,195}

\begin{axis}[
width=1\linewidth,
height=3.5cm,
legend cell align={left},
legend style={fill opacity=0.8, draw opacity=1, text opacity=1, draw=lightgray204},
tick align=outside,
tick pos=left,
x grid style={darkgray176},
xmajorgrids,
xmin=-5, xmax=105,
xtick style={color=black},
y grid style={darkgray176},
ylabel={Mean cost},
ymajorgrids,
ymin=-44.9076851443811, ymax=12.4224212332205,
ytick style={color=black}
]
\path [draw=red, fill=red, opacity=0.2]
(axis cs:0,9.42714232206345)
--(axis cs:0,9.42714232206345)
--(axis cs:24,-40.6767906275662)
--(axis cs:27,-40.9366134730252)
--(axis cs:33,-42.3017712181265)
--(axis cs:33,-38.2869800383394)
--(axis cs:33,-38.2869800383394)
--(axis cs:27,-34.7231390367855)
--(axis cs:24,-36.8954822366888)
--(axis cs:0,9.42714232206345)
--cycle;

\path [draw=mediumturquoise64181195, fill=mediumturquoise64181195, opacity=0.2]
(axis cs:0,9.42714165015654)
--(axis cs:0,9.42714165015654)
--(axis cs:25,-28.4057653546333)
--(axis cs:50,-37.7336309985681)
--(axis cs:75,-39.7950520678)
--(axis cs:100,-40.1434959823435)
--(axis cs:100,-39.7950526259162)
--(axis cs:100,-39.7950526259162)
--(axis cs:75,-37.7336326078935)
--(axis cs:50,-33.2677841620012)
--(axis cs:25,9.81650730696592)
--(axis cs:0,9.42714165015654)
--cycle;

\addplot [semithick, red, mark=*, mark size=1, mark options={solid}]
table {%
0 9.42714232206345
24 -38.8168956799941
27 -35.9598205306313
33 -40.7187514359301
};
\addlegendentry{CACTO-BIC}
\addplot [semithick, mediumturquoise64181195, mark=*, mark size=1, mark options={solid}]
table {%
0 9.42714165015654
25 -18.5943375934254
50 -34.3337478150021
75 -39.7950518022884
100 -39.7950527830557
};
\addlegendentry{PPO (Brax)}
\end{axis}

\end{tikzpicture}

%% file: Fig/CvsB_M_HR.tex
\begin{tikzpicture}

\definecolor{darkgray176}{RGB}{176,176,176}
\definecolor{lightgray204}{RGB}{204,204,204}
\definecolor{mediumturquoise64181195}{RGB}{64,181,195}

\begin{axis}[
width=1\linewidth,
height=3.5cm,
legend cell align={left},
legend style={fill opacity=0.8, draw opacity=1, text opacity=1, draw=lightgray204},
tick align=outside,
tick pos=left,
x grid style={darkgray176},
xlabel={Computation time [s]},
xmajorgrids,
xmin=-1314, xmax=27594,
xtick style={color=black},
y grid style={darkgray176},
ylabel={Mean cost},
ymajorgrids,
ymin=-5.71122572370099, ymax=33.2307009448834,
ytick style={color=black},
extra y ticks={-10},
extra y tick labels={}
]
\path [draw=red, fill=red, opacity=0.2]
(axis cs:0,31.4606133690386)
--(axis cs:0,31.4606133690386)
--(axis cs:310,5.1916230343508)
--(axis cs:566,2.37806734826529)
--(axis cs:825,-1.38536129620942)
--(axis cs:1087,-1.64995936326908)
--(axis cs:1348,-0.0729762333812136)
--(axis cs:1606,-3.94113814785625)
--(axis cs:1606,-2.77041150403745)
--(axis cs:1606,-2.77041150403745)
--(axis cs:1348,4.00173153705669)
--(axis cs:1087,5.15249805829742)
--(axis cs:825,3.22300220032533)
--(axis cs:566,6.95261193540963)
--(axis cs:310,11.3364889915242)
--(axis cs:0,31.4606133690386)
--cycle;

\path [draw=mediumturquoise64181195, fill=mediumturquoise64181195, opacity=0.2]
(axis cs:0,31.4606133690386)
--(axis cs:0,31.4606133690386)
--(axis cs:2190,11.0470459495078)
--(axis cs:4380,4.12402176766685)
--(axis cs:6570,1.57131174500241)
--(axis cs:8760,3.16229937203003)
--(axis cs:10950,3.24582834744995)
--(axis cs:13140,2.63188435769442)
--(axis cs:15330,1.22886560925029)
--(axis cs:17520,3.02280230639559)
--(axis cs:19710,2.51412205623858)
--(axis cs:21900,2.68653213526263)
--(axis cs:24090,0.909411784600128)
--(axis cs:26280,0.261535876628124)
--(axis cs:26280,2.76105042321212)
--(axis cs:26280,2.76105042321212)
--(axis cs:24090,4.57313837652857)
--(axis cs:21900,4.49901169290145)
--(axis cs:19710,4.53550347821279)
--(axis cs:17520,3.56531452364994)
--(axis cs:15330,2.75481745826476)
--(axis cs:13140,6.12112113246412)
--(axis cs:10950,3.86243459446864)
--(axis cs:8760,4.45956850277655)
--(axis cs:6570,4.41362011432648)
--(axis cs:4380,8.88272495432333)
--(axis cs:2190,13.8914646002831)
--(axis cs:0,31.4606133690386)
--cycle;

\addplot [semithick, red, mark=*, mark size=1, mark options={solid}]
table {%
0 31.4606133690386
310 6.94999563219872
566 2.67567512618773
825 -1.04122739578738
1087 -1.14456304727179
1348 1.18177084082907
1606 -2.96313725682822
};
\addlegendentry{CACTO-BIC}
\addplot [semithick, mediumturquoise64181195, mark=*, mark size=1, mark options={solid}]
table {%
0 31.4606133690386
2190 11.5046886919123
4380 7.15270161086863
6570 2.6485093213392
8760 3.63858486350739
10950 3.75912724299864
13140 3.18334943101262
15330 2.41537656589891
17520 3.13646657019854
19710 3.2237353135239
21900 3.4012841166872
24090 1.10466703230684
26280 2.11331907321106
};
\addlegendentry{PPO (Brax)}
\end{axis}

\end{tikzpicture}

%% file: Fig/CvsB_R.tex
\begin{tikzpicture}

\definecolor{darkgray176}{RGB}{176,176,176}
\definecolor{lightgray204}{RGB}{204,204,204}
\definecolor{mediumturquoise64181195}{RGB}{64,181,195}

\begin{axis}[
width=1\linewidth,
height=4cm,
legend cell align={left},
legend style={fill opacity=0.8, draw opacity=1, text opacity=1, draw=lightgray204},
tick align=outside,
tick pos=left,
x grid style={darkgray176},
xlabel={Computation time [s]},
xmajorgrids,
xmin=-21.25, xmax=446.25,
xtick style={color=black},
y grid style={darkgray176},
ylabel={Mean cost},
ymajorgrids,
ymin=3.40855257433114, ymax=7.2869940333892,
ytick style={color=black}
]
\path [draw=red, fill=red, opacity=0.2]
(axis cs:0,7.11070123979565)
--(axis cs:0,7.11070123979565)
--(axis cs:20,3.93947549970801)
--(axis cs:28,3.72183690078019)
--(axis cs:35,3.60484618262493)
--(axis cs:44,3.59450834686148)
--(axis cs:53,3.58484536792468)
--(axis cs:53,3.80730473528587)
--(axis cs:53,3.80730473528587)
--(axis cs:44,3.78389256750954)
--(axis cs:35,3.82908311670588)
--(axis cs:28,3.93902708309674)
--(axis cs:20,4.24459492695547)
--(axis cs:0,7.11070123979565)
--cycle;

\path [draw=mediumturquoise64181195, fill=mediumturquoise64181195, opacity=0.2]
(axis cs:0,7.1007825552458)
--(axis cs:0,7.1007825552458)
--(axis cs:25,5.24787769305337)
--(axis cs:50,4.90783479514042)
--(axis cs:75,4.88570607783868)
--(axis cs:100,4.76197062021832)
--(axis cs:125,4.64718192923315)
--(axis cs:150,4.48983881011913)
--(axis cs:175,4.22241176985992)
--(axis cs:200,4.18139133000862)
--(axis cs:225,4.12570956599966)
--(axis cs:250,4.18931057259978)
--(axis cs:275,4.11738339744835)
--(axis cs:300,4.07259014858528)
--(axis cs:325,4.0410030540679)
--(axis cs:350,4.02902308824597)
--(axis cs:375,4.00444662423574)
--(axis cs:400,3.96679174086228)
--(axis cs:425,3.96686804604827)
--(axis cs:425,4.0489488405958)
--(axis cs:425,4.0489488405958)
--(axis cs:400,4.08709562501088)
--(axis cs:375,4.13359859658732)
--(axis cs:350,4.09742427207818)
--(axis cs:325,4.16892922016643)
--(axis cs:300,6.06206233138891)
--(axis cs:275,4.17802401492132)
--(axis cs:250,4.24478111854879)
--(axis cs:225,4.32144742214947)
--(axis cs:200,4.35685951407692)
--(axis cs:175,4.46451149174233)
--(axis cs:150,4.66529162712011)
--(axis cs:125,5.29483222531289)
--(axis cs:100,4.84648962557009)
--(axis cs:75,5.21801112937866)
--(axis cs:50,5.31286238514117)
--(axis cs:25,5.68780432641506)
--(axis cs:0,7.1007825552458)
--cycle;

\addplot [semithick, red, mark=*, mark size=1, mark options={solid}]
table {%
0 7.11070123979565
20 3.94805790745106
28 3.7847241665976
35 3.67023963383255
44 3.60836617721571
53 3.6000087459466
};
\addlegendentry{CACTO-BIC}
\addplot [semithick, mediumturquoise64181195, mark=*, mark size=1, mark options={solid}]
table {%
0 7.1007825552458
25 5.33715863971366
50 4.99792725861687
75 5.20179238443061
100 4.78738980210319
125 4.70597913869916
150 4.50018688596631
175 4.41318263704697
200 4.35677532095116
225 4.30411391540133
250 4.20032612439821
275 4.13529680174843
300 4.07932564037427
325 4.09882224254915
350 4.04958991412813
375 4.05342334056886
400 4.01826757149676
425 3.9965366263557
};
\addlegendentry{PPO (Brax)}
\end{axis}

\end{tikzpicture}